\documentclass[5pt]{article}

\usepackage[letterpaper]{geometry}
\usepackage{enumitem}
\usepackage{spconf,amsmath,amssymb, amsfonts}
\usepackage{amsthm}
\usepackage{url}
\usepackage{bbm}
\usepackage{graphicx}
\usepackage{subcaption}
\usepackage{caption}
\usepackage{float}
\usepackage{tikz}
\usetikzlibrary{shapes.geometric, arrows,positioning}

\newtheorem{lemma}{Lemma}[section]
\newtheorem{theorem}{Theorem}[section]

\usepackage{fancyhdr}
\begin{document}\sloppy

\def\x{{\mathbf x}}
\def\L{{\cal L}}

\newcommand{\etal}{\textit{et al}. }

\def\exp{{\mathrm{exp}}}

\title{Confidence estimation in Deep Neural networks via density modelling}
%
\name{Akshayvarun Subramanya~~~~~~ Suraj Srinivas~~~~~~ R.Venkatesh Babu
}
\address{Video Analytics Lab, Department of Computational and Data Sciences\\ Indian Institute of Science, Bangalore\\
\texttt{\small akshayvarun07@gmail.com}, \texttt{\small surajsrinivas@grads.cds.iisc.ac.in}, \texttt{\small venky@cds.iisc.ac.in}}
%
%

\maketitle

\begin{abstract}
State-of-the-art Deep Neural Networks can be easily fooled into providing incorrect high-confidence predictions for images with small amounts of adversarial noise. Does this expose a flaw with deep neural networks, or do we simply need a better way to estimate confidence? In this paper we consider the problem of accurately estimating predictive confidence. We formulate this problem as that of density modelling, and show how traditional methods such as softmax produce poor estimates. To address this issue, we propose a novel confidence measure based on density modelling approaches. We test these measures on images distorted by blur, JPEG compression, random noise and adversarial noise. Experiments show that our confidence measure consistently shows reduced confidence scores in the presence of such distortions - a property which softmax often lacks.
\end{abstract}
\section{Introduction}
\label{sec:intro}
Deep neural networks have contributed to tremendous advances in Computer Vision during recent times \cite{krizhevsky2012imagenet, Simonyan15,he2015deep}. For classification tasks, the general practice has been to apply a \textit{softmax} function to the network's output. The main objective of this function is to produce a probability distribution over labels such that most of the mass is situated at the maximum entry of the output vector. While this is essential for training, softmax is often retained at test time, and the output of this function is often interpreted as an estimate of the true underlying distribution over labels given the image.

Images can often be corrupted by artifacts such as random noise and filtering. We require classifiers to be robust to such distortions. Recently, Goodfellow \etal \cite{goodfellow2014explaining} showed that it is possible for an adversary to imperceptibly change an image leading to high-confidence false predictions. This places Deep Neural Networks at a severe disadvantage when it comes to applications in forensics or biometrics.

Recent works \cite{Gal2015DropoutB, bendale2016towards} have empirically demonstrated that the softmax function is often ineffective at producing accurate uncertainty estimates. By producing better estimates, is it possible to detect such adversarial examples? This leads us to ask - what constitutes a good uncertainty / confidence estimate? Is there a fundamental flaw in estimating confidences using softmax? This paper discusses these issues and proposes a novel density modelling-based solution to this end. The overall contributions of this paper are:

\begin{itemize}
    \item We discuss the general problem of uncertainty / confidence estimation and show how softmax can exhibit pathological behaviour.
    \item We propose a novel method for estimating predictive confidence based on density modelling.
    \item We provide experimental evidence showing that the proposed method is indeed superior to softmax at producing confidence estimates.
\end{itemize}

This paper is organized as follows. Section 2 describes different approaches that have been taken to tackle the confidence estimation problem. Section 3 introduces terminology and describes our approach. Section 4 describes experimental setup and results. Finally, in Section 5 we present our conclusions.

\section{Related works}
Uncertainty or Confidence estimation has gained a lot of attention in recent times. Gal \etal \cite{Gal2015DropoutB} presented a method of estimating the uncertainty in neural network model by performing dropout averaging of predictions during test time. Bendale \etal \cite{bendale2016towards} presented Open set deep networks, which attempt to determine whether a given image belongs to any of the classes it was trained for. However, both these methods use the softmax function to compute uncertainty. We will show shortly that uncertainty estimates from softmax contain certain pathologies, making them unsuitable for this task.

Modern neural network architectures are sensitive to adversarial examples \cite{szegedy2013intriguing}. These are images produced by the addition of small perturbations to correctly classified samples. Generating adversarial examples to \textit{fool} the classifier is one of the active areas of research in the Deep Learning Community. Goodfellow  \etal \cite{goodfellow2014explaining} presented a method to generate adversarial examples and also showed that retraining the network with these examples can be used for regularization. Nyugen \etal\cite{nguyen2015deep} also strengthened the claim that networks can be fooled easily by generating \textit{fooling images}. Moosavi \etal \cite{moosavi2016deepfool} presented an effective and fast way of generating the perturbations required to misclassify any image. Moosavi \etal \cite{moosavi2016universal} also show that there exists universal adversarial perturbations given a classifier, that can be applied to any image for misclassification . Researchers have also shown that results of face recognition algorithms can be tampered by wearing a specific design of eyeglass frames \cite{Sharif16AdvML}. Such examples present a challenge to the idea of using neural networks for commercial applications, since security could be easily compromised. These can be overcome (in principle) by using a good uncertainty estimate of predictions. 

 \section{Confidence estimation}
We first introduce the concept of predictive confidence in a neural network in a classification setting. Let $(X,y)$ be random variables representing the training data where $X \in \Re^D, y \in \Re^N$  and $f(X):\Re^{D}\rightarrow \Re^{N}$ be a function that represents the mapping of $D$-dimensional input data to the pre-softmax layer ($z$) of $N$ dimensions, where $N$ is the number of output classes. In this case, we define confidence estimation for a given input $X$ as that of estimating $P(y|X)$ from $z$. 

Intuitively, confidence estimates are also closely related to accuracy. While accuracy is a measure of how well the classifier's outputs align with the ground truth, confidence is the model's estimate of accuracy in absence of ground truth. Ideally we would like our estimates to be correlated with accuracy, i.e high accuracy $\Rightarrow$ high confidence on average, for a given set of samples. A model with low accuracy and high confidence indicates that the model is very confident about making incorrect predictions, which is undesirable.

\begin{figure}[!b] 
  \label{fig:softmax_path} 
  \begin{minipage}[b]{\linewidth}
    \centering
    \begin{subfigure}[b]{2.8cm}
    \includegraphics[width=2.8cm]{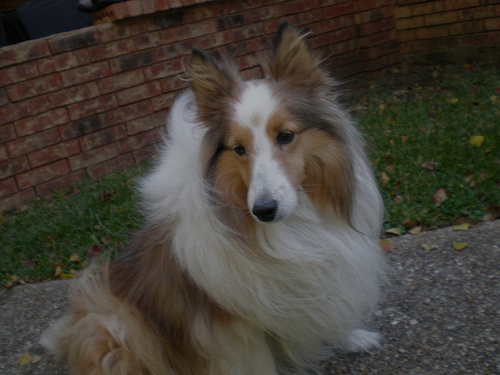}
    \caption{0.7 $\times$ Image\\\textbf{Softmax}: 0.8791 }\label{fig:soft_path1}
    \end{subfigure}
    \begin{subfigure}[b]{2.8cm}
    \includegraphics[width=2.8cm]{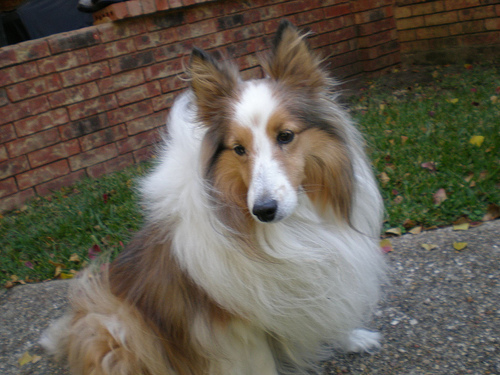}
    \caption{Original Image\\ \textbf{Softmax}: 0.9249 }\label{fig:soft_path2}
    \end{subfigure}
    \begin{subfigure}[b]{2.8cm}
    \includegraphics[width=2.8cm]{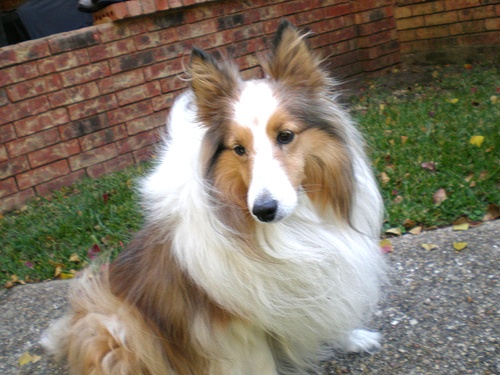}
    \caption{1.3 $\times$ Image\\ \textbf{Softmax}: 0.9687 }\label{fig:soft_path3}
    \end{subfigure}
  \end{minipage}
  \caption{An illustration of the softmax pathology on an image from the ImageNet dataset using the VGG-16 classifier.}
  \label{fig:pathology}
  \end{figure}

\begin{figure*}[!ht]
\begin{subfigure}{.5\textwidth}
  \centering
\includegraphics[width = 7.7cm]{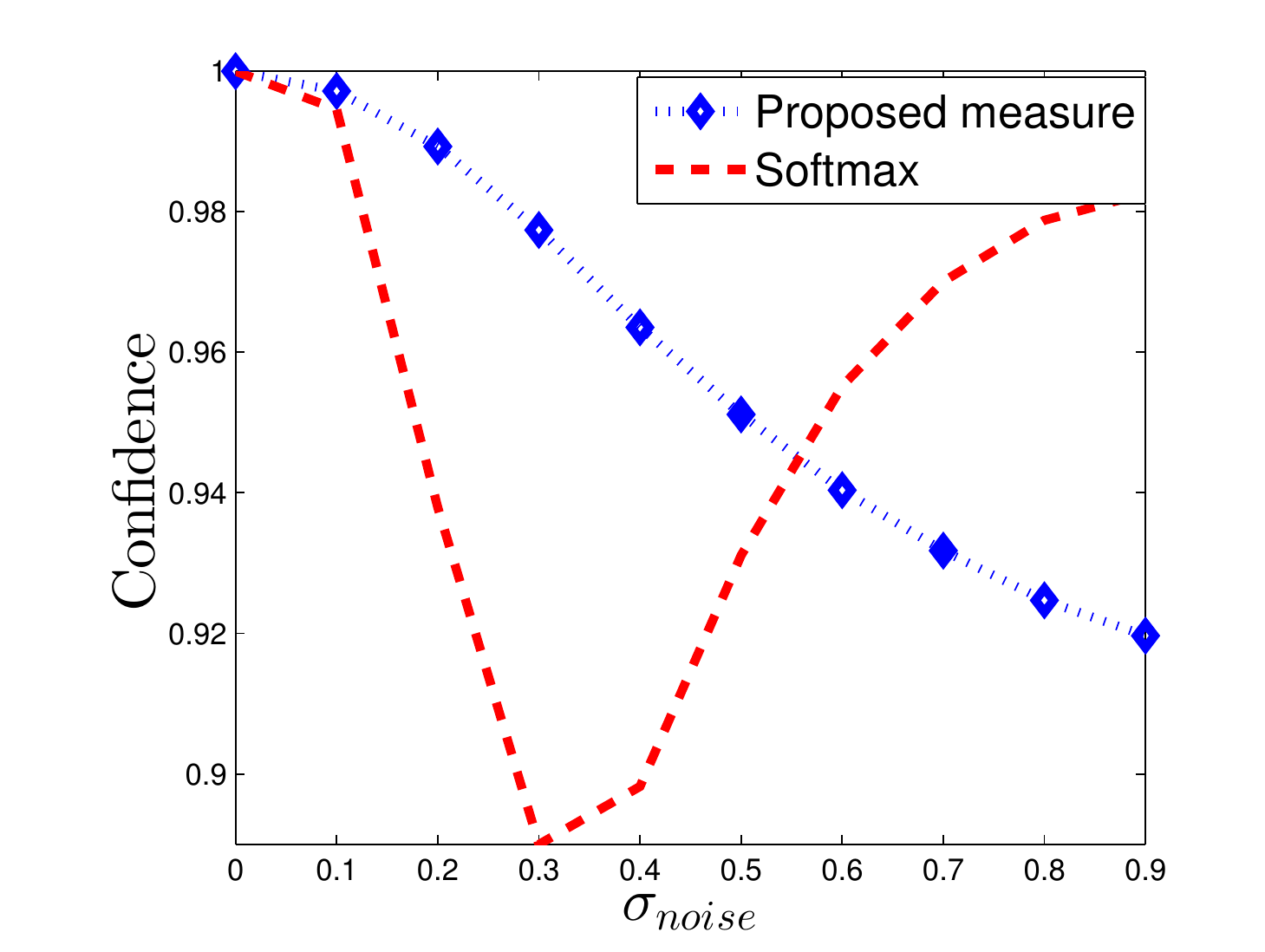}
\caption{{Gaussian Noise}}
\label{fig:mnist_dist1}
\end{subfigure} 
\begin{subfigure}{.5\textwidth}
  \centering
\includegraphics[width = 7.7cm]{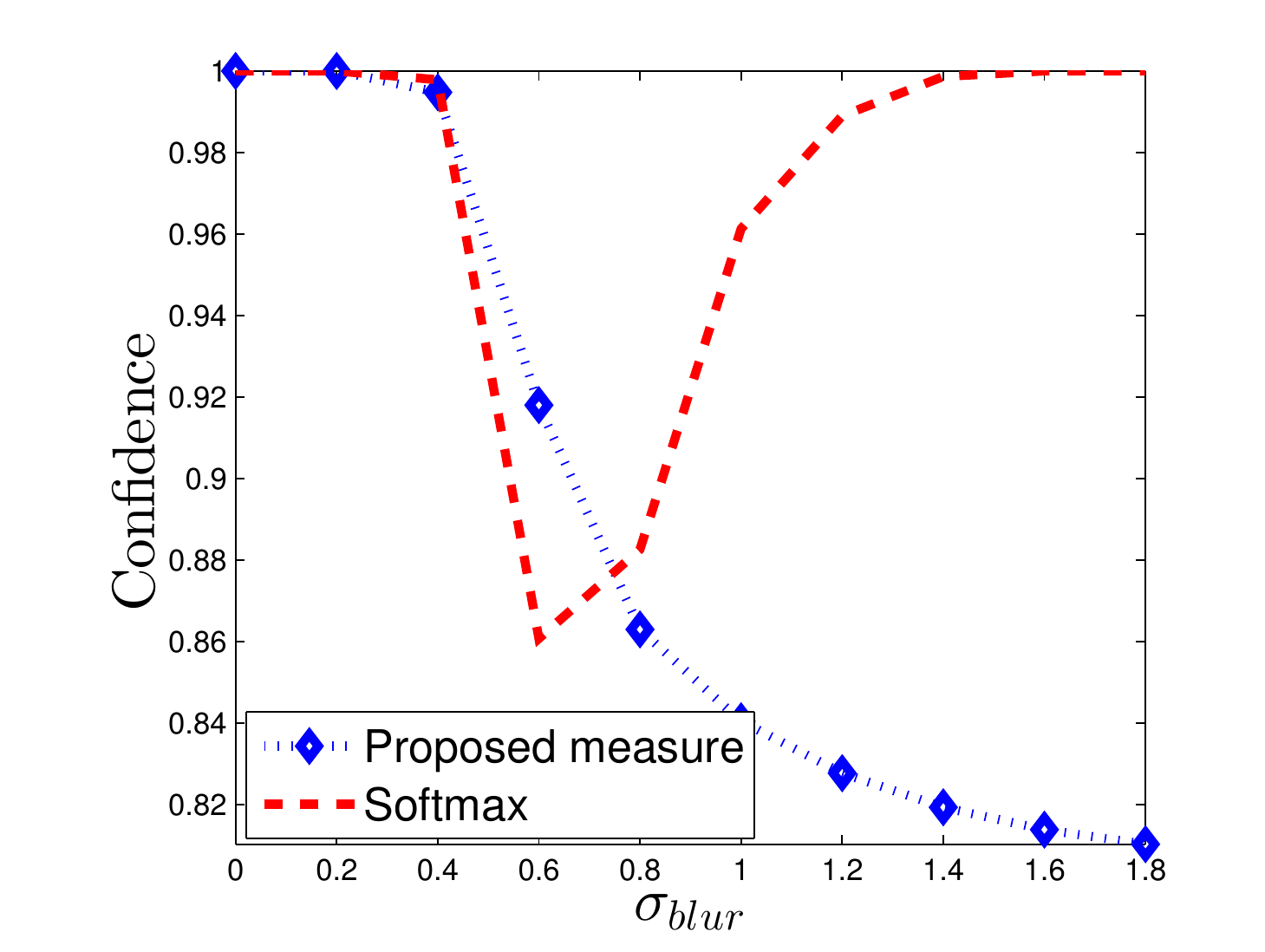}
\caption{{Gaussian Blurring}}
\label{fig:mnist_dist2}
\end{subfigure}%
\vspace{-1pt}
\caption{Comparison of confidence measures for different distortions applied on MNIST images. A good confidence measure must exhibit a monotonically decreasing profile. We see that the proposed method does indeed have such a profile, whereas softmax shows high confidence even for very high distortions. Note that here both confidence measures are scaled such that clean images correspond to confidence of one, for better visualization.}
\label{fig:conf_qual_mnist}
\end{figure*}

\subsection{Pathologies with Softmax and Neural Networks}
Given the definitions above, and pre-softmax activation vector with elements $z = [z_1, z_2, ... z_N]$, the softmax function is defined as follows.

\begin{equation}
P_s(y_i|X) = s_i(z) = \frac{e^{z_i}}{\sum_i e^{z_i} }
\label{eq:softmax}
\end{equation}
Here, $P_s(y_i|
X)$ denotes the softmax estimate of the desired probability $P(y_i|X)$ for label $y_i$. We can easily see that for neural networks with monotonically increasing activation functions, $f(k X) \geq f(X)$, for any $k > 1$. This is because this property of linear scaling applies to all layers of a neural network - convolutions, fully connected layers, max-pooling, batch normalization and commonly used activation functions. As a result, it applies to the entire neural network as a whole. Hence, the pre-softmax vector transforms to $\|z'\| = \| f(k X) \| \geq \|z \|$. This also trivially holds for  multi-class linear classifiers of the form $g(X) = W^T X$. In such cases, the following lemma applies.

\begin{lemma}
Let $z = [z_1, z_2, ... z_i, ...z_N]$ be a pre-softmax activation vector, such that $z_i = max(z_1, ... z_N)$. Given a positive scalar $k > 1$ and the softmax activation function $s_i(z)$ given in Equation (\ref{eq:softmax}), the following statement is always true

\begin{equation*}
    s_i(k z) > s_i(z)
\end{equation*}

\label{lemma}
\end{lemma}
The proof for this lemma appears in the Appendix. This implies that for softmax, $P_s(y_i|k X) > P_s(y_i|X)$. This also indicates that irrespective of the structure of data, an input $X$ with large $\ell_2$ norm always produces higher confidence than that with a lower $\ell_2$ norm. This exposes a simple way to boost confidence for any image - by simply increasing the magnitude. This is illustrated with an example in Figure \ref{fig:pathology}. Clearly, this method of computing confidence has pathologies and therefore, must be avoided.

\subsection{How to estimate better confidence?} 
The traditional notion of confidence for linear classifiers relies on the concept of distance from the separating hyper-plane, i.e.; the farther the point is from the hyper-plane, the more confident we are of the point belonging to one class \cite{platt1999probabilistic}. However, this reasoning is fundamentally flawed - points with large $\ell_2$ norms are more likely to lie far away from a given hyper-plane. As a result, such points are always in one class or another with very high confidence. This clearly ignores the structure and properties of the specific dataset in question. 

A much better notion of confidence would be to measure distances with points of either class. If a given point is closer to points of one class than another, then it is more likely to fall in that class. Points at infinity are at a similar distance from points of all classes. This clearly provides a much better way to estimate confidence. We shall now look at ways to formalize this intuition.  
 
\subsection{Proposed method: Density Modelling}
According to the intuition presented above, we require to characterize the set of all points belonging to a class. The most natural way to do that is to create a density model of points in each class. As a result, for each class $y_i$, we compute $P(z|y_i)$, where $z$ is the activation of the final deep layer, previously referred to as the pre-softmax activation. Given these density models, the most natural way to obtain $P(y_i|z)$ is to use Bayes Rule.

\begin{equation}
    P(y_i|z) = \frac{P(z|y_i) P(y_i)}{\sum \limits_{j=1}^{N} P(z|y_i) P(y_i)}
\label{eqn:bayes}
\end{equation}

This lets us compute $P(y_i|z)$ efficiently. However we mentioned that we wish to compute $P(y_i|X)$ rather than $P(y_i|z)$. Since the mapping from $X$ to $z$ is deterministic (given by a neural network), we assume that $P(z) \sim P(X)$. Although the validity of this assumption may be debatable, we empirically know that there exists a one-to-one mapping from a large family of input images $X$ to corresponding features $z$. In other words, it is extremely rare to have two different natural images with exactly the same feature vector $z$. This assumption empirically seems to hold for a large class of natural images. Here, the prior $P(y_i)$ is based on the distribution of classes in training data. 

In this work, we perform density modelling using multivariate Gaussian densities with a diagonal covariance matrix. Note that this is not a limitation of our method - the assumptions only make computations simple. As a result, if there are $N$ classes in a classification problem, we compute parameters of $N$ such Gaussian densities ($\mu, \sigma$).

\begin{equation}
    P(X|y_i) = \mathcal{N}(z | \mu_i, \sigma_i)
\label{eqn:normal}
\end{equation}

After we evaluate the likelihood for each class, we apply Bayes rule i.e Equation (\ref{eqn:bayes}) by multiplying the prior and then normalising it, giving rise to confidence measure. 

\subsection{Gaussian in High Dimensions}
High dimensional Gaussian densities are qualitatively different from low-dimensional ones. The following theorem explains this phenomenon.

\begin{theorem}[Gaussian Annulus Theorem]
\label{GaussAnnulus}
For a d-dimensional spherical Gaussian with unit variance in each direction, for any $\beta  \leq \sqrt{d}$, all but at most $3e^{-c\beta^{2}}$ of the probability mass lies within the annulus $\sqrt{d} - \beta \leq | x |  \leq \sqrt{d} + \beta$  where c is a fixed positive coefficient.
\end{theorem}

The proof for this theorem can be found in \cite{hopcroft2014foundations}. This theorem implies that for a high-dimensional Gaussian with unit variance, nearly all of the probability is concentrated in a thin annulus of width $O(1)$ with mean distance of $\sqrt{d}$ from the centre. This implies that almost all points within that density have more or less the same vanishingly small probability. This presents a problem when performing computations using Bayes rule. We require density functions such that different points have vastly differing densities.

One way to overcome this problem is to compute densities using a covariance of $d \times \sigma^2$ instead of $\sigma^2$. This ensures that majority of the points fall around the covariance rather than farther away. The resulting density values show variation among points, and do not have vanishingly small values, unlike in the previous case. 

\subsection{Overall process}
Here we describe our confidence estimation process, and how to obtain confidence for a given new image. Training is performed as usual, using the softmax activation function. After training, the training data is re-used to calculate the parameters of the density distribution in Equation \ref{eqn:normal}. At test time, the label is obtained as before - by looking at the maximum entry of $z$ (which is the same as the maximum entry of softmax output). However, confidence is obtained by first calculating all $N$ density values and then applying Bayes' rule (Equation \ref{eqn:bayes}).

\begin{figure}[!hb] 
  \label{fig:mnist_illustartion} 
  \begin{minipage}[b]{\linewidth}
    \centering
    \begin{subfigure}[b]{2.5cm}
    \includegraphics[width=2.5cm]{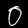}
    \caption{Clean Image\\\textbf{Softmax}: 0.999 \\\textbf{Ours} : 0.957}\label{fig:mnist_noise1}
    \end{subfigure}
    \begin{subfigure}[b]{2.5cm}
    \includegraphics[width=2.5cm]{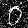}
    \caption{Noise ($\sigma$=0.3)\\ \textbf{Softmax}: 0.921 \\\textbf{Ours}: 0.751}\label{fig:mnist_noise2}
    \end{subfigure}
    \begin{subfigure}[b]{2.5cm}
    \includegraphics[width=2.5cm]{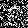}
    \caption{Noise (\textbf{$\sigma$}=0.8)\\ \textbf{Softmax}: 0.994 \\\textbf{Ours}: 0.57 }\label{fig:mnist_noise3}
    \end{subfigure}
  \end{minipage}
  
   \begin{minipage}[b]{\linewidth}
    \centering
    \begin{subfigure}[b]{2.5cm}
    \includegraphics[width=2.5cm]{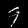}
    \caption{Clean Image\\\textbf{Softmax}: 0.388 \\\textbf{Ours}: 0.961}\label{fig:mnist_blur1}
    \end{subfigure}
    \begin{subfigure}[b]{2.5cm}
    \includegraphics[width=2.5cm]{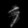}
    \caption{Blur (\textbf{$\sigma$}=0.6)\\ \textbf{Softmax}: 0.667 \\\textbf{Ours}: 0.566}\label{fig:mnist_blur2}
    \end{subfigure}
    \begin{subfigure}[b]{2.5cm}
    \includegraphics[width=2.5cm]{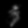}
    \caption{Blur (\textbf{$\sigma$}=1.2)\\ \textbf{Softmax}: 0.807 \\\textbf{Ours}: 0.287}\label{fig:mnist_blur3}
    \end{subfigure}
  \end{minipage}
  
  \caption{An illustration of the effectiveness of our method on MNIST. The proposed confidence measure decreases when distortions are added to the image, while softmax remains high.}
\end{figure}

\begin{figure*}[!ht]
\begin{subfigure}{.33\textwidth}
  \centering
\includegraphics[width = 6.4cm]{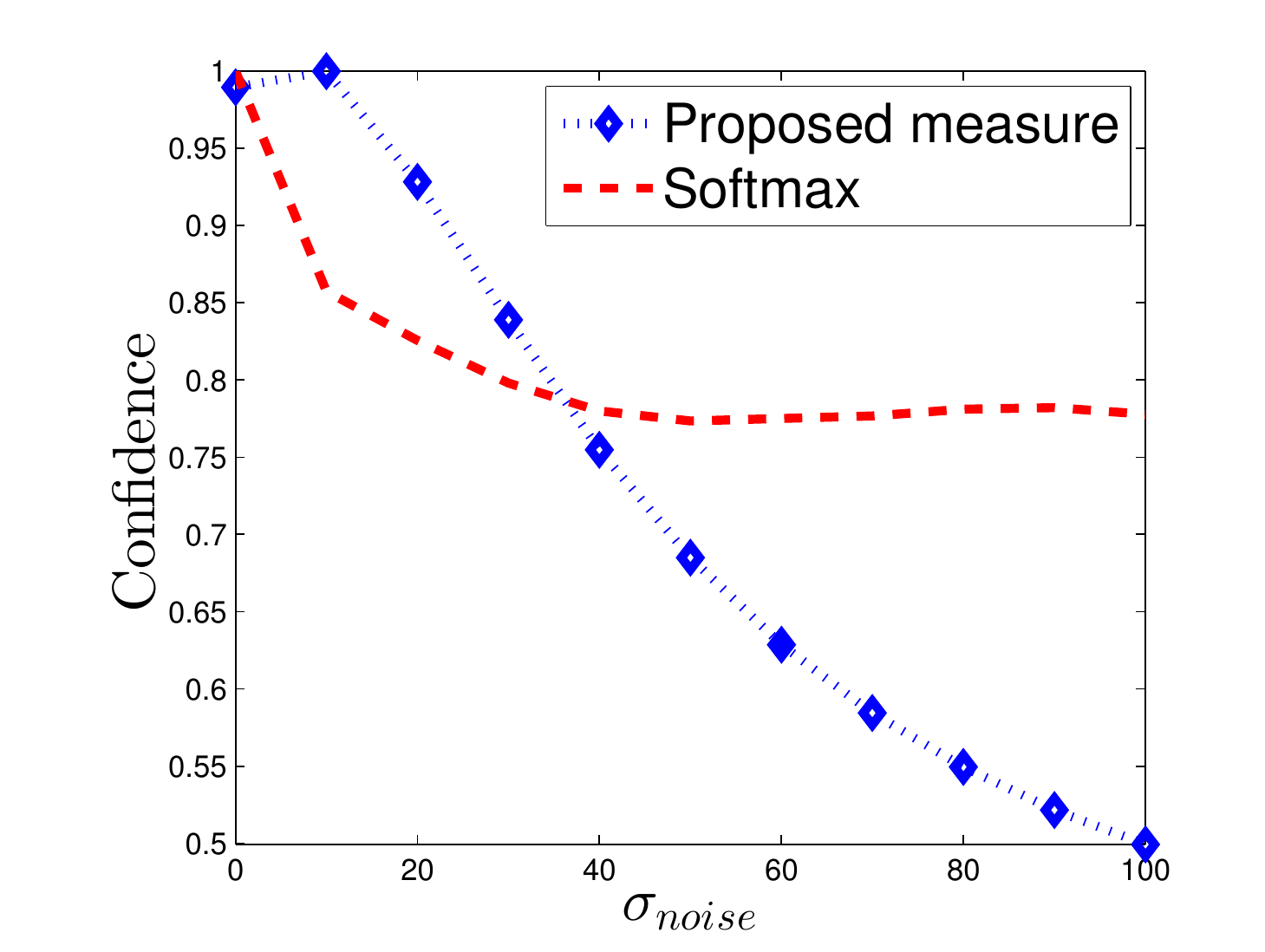}
\caption{{Gaussian Noise}}
\label{fig:dist1}
\end{subfigure} 
\begin{subfigure}{.33\textwidth}
  \centering
\includegraphics[width = 6.4cm]{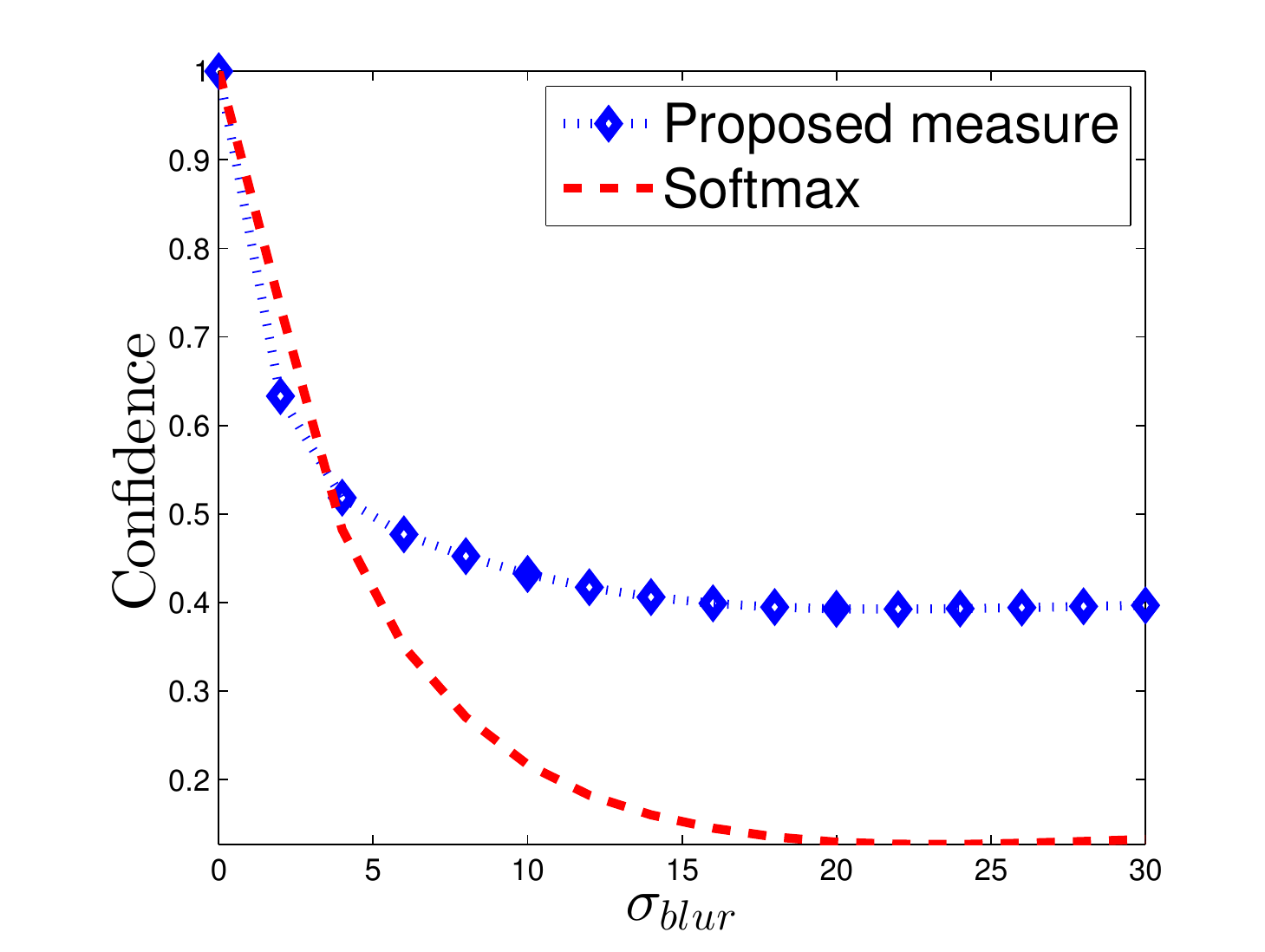}
\caption{{Gaussian Blurring}}
\label{fig:dist2}
\end{subfigure}%
\begin{subfigure}{.33\textwidth}
  \centering
\includegraphics[width = 6.4cm]{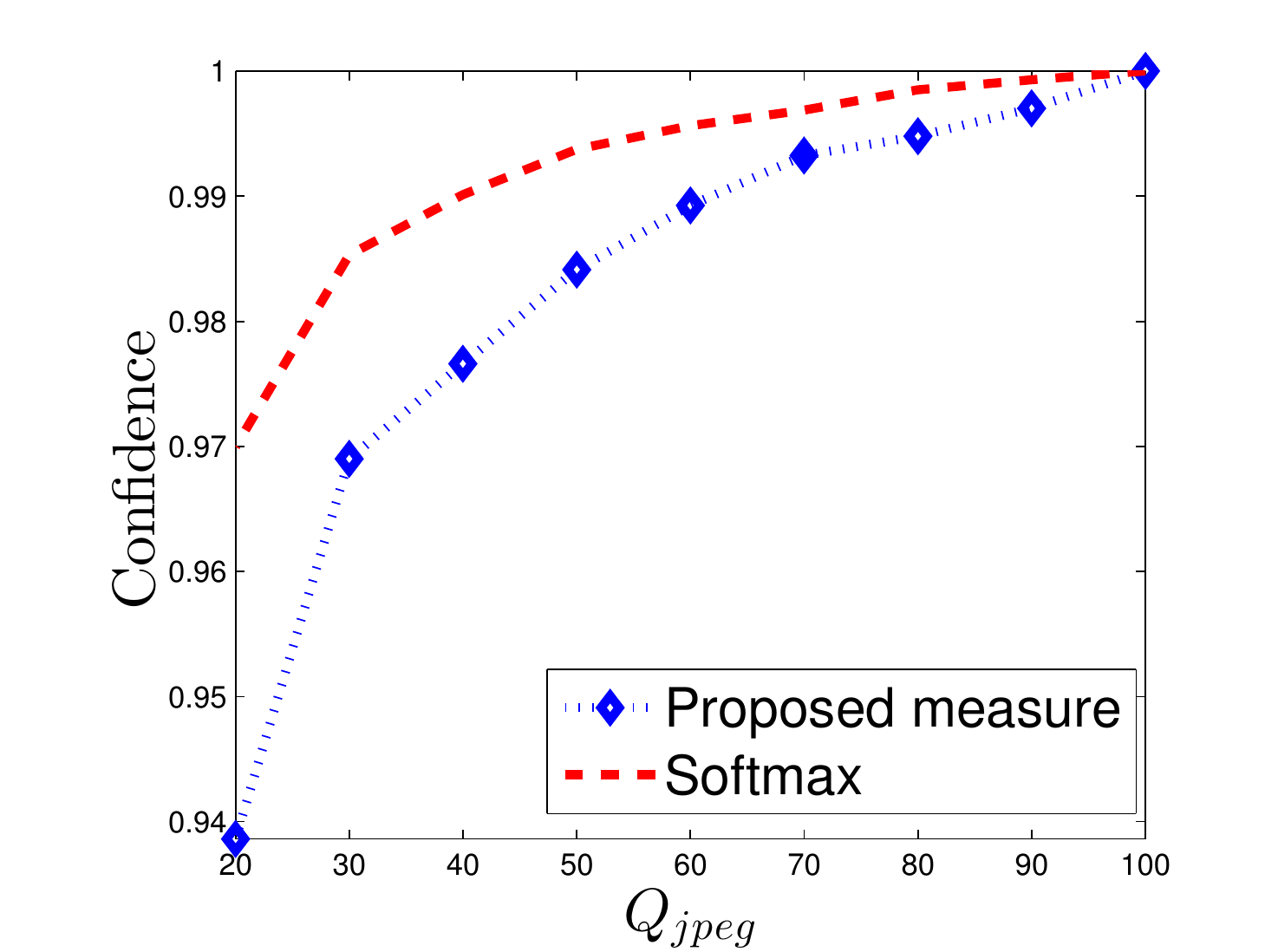}
\caption{{JPEG Compression}}
\label{fig:dist3}
\end{subfigure}%

\caption{Comparison of confidence measures for various types of distortions applied on images in the ImageNet dataset. Both confidence measures are scaled such that the clean image always obtains confidence value = 1. This is done for better visualization. Figure (a) shows that the proposed approach is qualitatively better then softmax, which does not have a monotonically decreasing profile. For Figures (b-c), the proposed approach and softmax behave similarly.}
\label{fig:conf_qual}
\end{figure*}

\section{Experiments}
We evaluate our method on two datasets - MNIST handwritten digit dataset  \cite{726791} and validation set of ILSVRC12 dataset \cite{ILSVRC15}. For the MNIST dataset, we consider the LeNet-5 architecture with 2 convolution layers and 2 fully connected layers. For ImageNet, we consider VGG-16 architecture \cite{Simonyan15}. 

When presented with an unfamiliar image such as those with different types of distortions, a good measure of confidence must present predictions with reduced confidences. Examples of distortions include Gaussian Blurring, Gaussian Noise, JPEG Compression, Thumbnail resizing similar to those considered in \cite{zheng2016improving}. In our experiments, we test this property of confidence measures on the following distortions.
\begin{itemize}
\item 
\textbf{Gaussian Noise:} Additive Noise drawn from a Gaussian distribution is one of the most common types of distortions that can occur in natural images. Here, we add incremental amounts of such noise and successively evaluate the confidence of the network. For MNIST, we vary the standard deviation of noise between 0 and 1, while for ImageNet it varies from 0 to 100. Note that the range of pixel values is [0,1] for MNIST and [0,255] for ImageNet.
We see that both for Figure 2(a) and Figure 4(a), our method exhibits the desired monotonically decreasing profile whereas softmax does not. 

\begin{figure*}[!ht] 
  
  \begin{minipage}[t]{0.5\linewidth}
    \centering
    \begin{subfigure}[t]{2.8cm}
    \includegraphics[width=2.8cm]{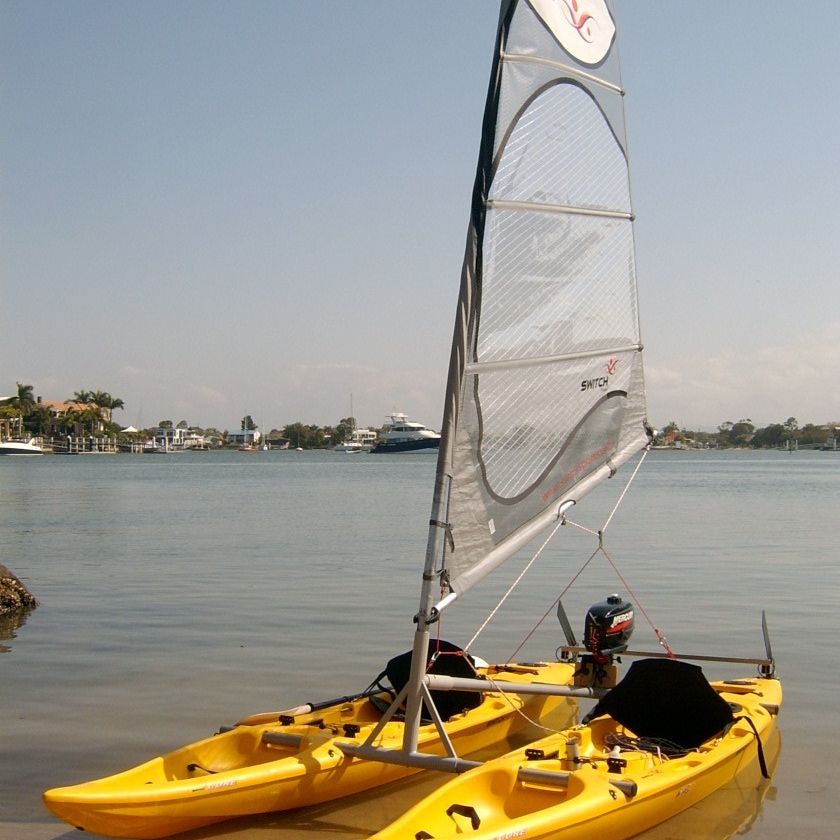}
    \caption{Clean image \\\textbf{Softmax}: 0.616\\\textbf{Ours}: 0.00302} \label{fig:noise_1}
    \end{subfigure}
    \begin{subfigure}[t]{2.8cm}
    \includegraphics[width=2.8cm]{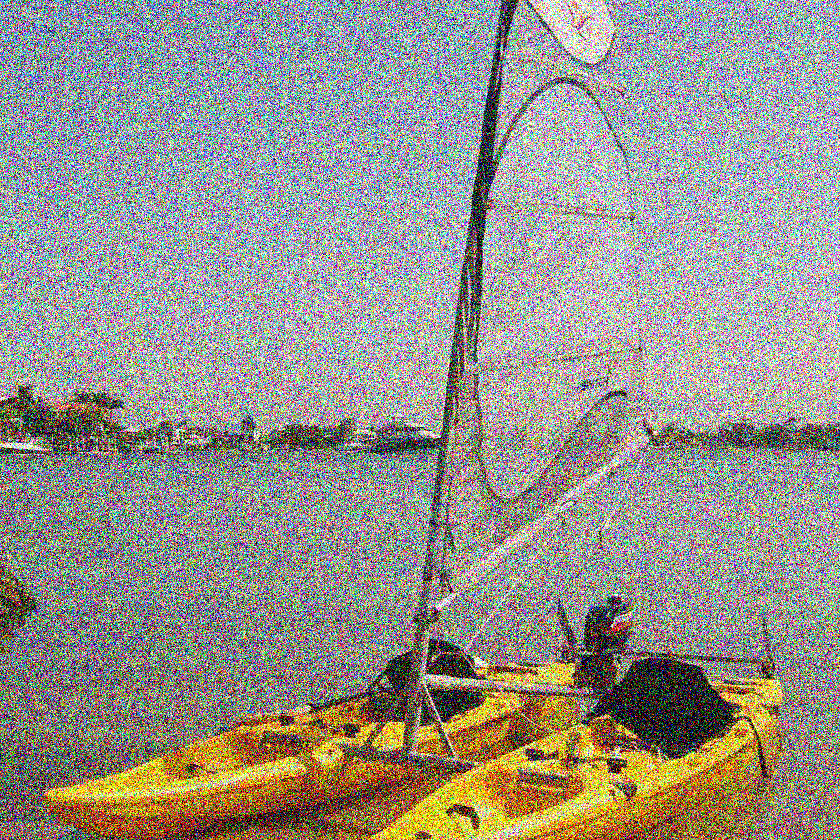}
    \caption{Noise ($\sigma = 50$)  \\\textbf{Softmax}: 0.620 \\\textbf{Ours}:  0.00236}\label{fig:noise_2}
    \end{subfigure}
    \begin{subfigure}[t]{2.8cm}
    \includegraphics[width=2.8cm]{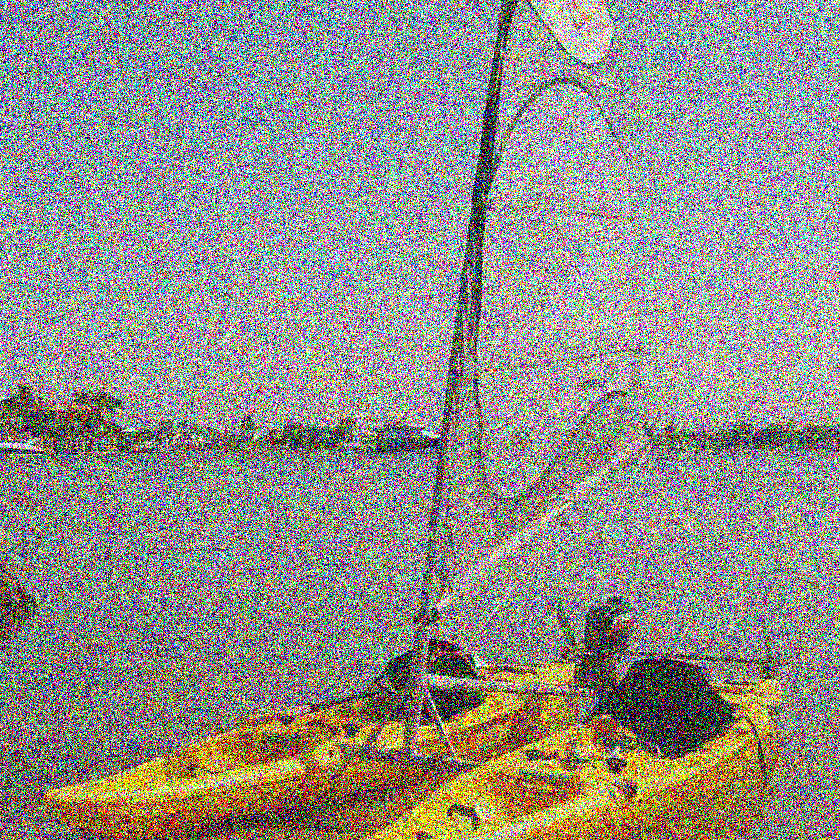}
    \caption{Noise ($\sigma = 100$) \\\textbf{Softmax}: 0.644 \\\textbf{Ours}: 0.00232}\label{fig:noise_3}
    \end{subfigure}
  \end{minipage}
  \begin{minipage}[t]{0.5\linewidth}
    \centering
    \begin{subfigure}[t]{2.8cm}
    \includegraphics[width=2.8cm]{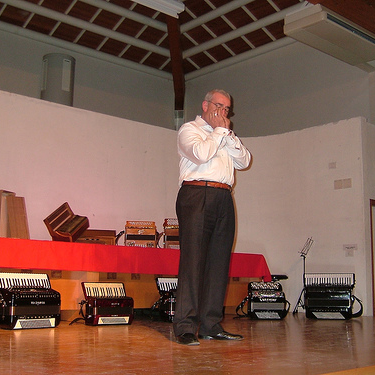}
    \caption{Clean image \\\textbf{Softmax}: 0.61628 \\\textbf{Ours}: 0.00302}
    \label{fig:blur_1}
    \end{subfigure}
    \begin{subfigure}[t]{2.8cm}
    \includegraphics[width=2.8cm]{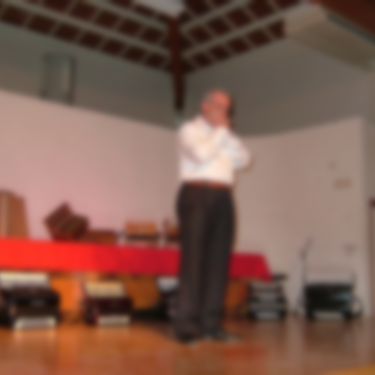}
    \caption{Blur ($\sigma = 3$) \\\textbf{Softmax}: 0.62077 \\\textbf{Ours}: 0.002367}
    \label{fig:blur_2}
    \end{subfigure}
    \begin{subfigure}[t]{2.8cm}
    \includegraphics[width=2.8cm]{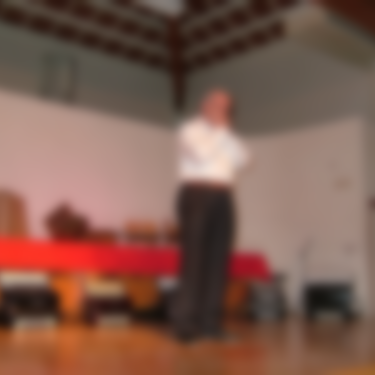}
    \caption{Blur ($\sigma = 5$) \\ \textbf{Softmax}: 0.64401 \\\textbf{Ours}: 0.002328}
    \label{fig:blur_3}
    \end{subfigure} 
  \end{minipage} 
  \begin{minipage}[b]{0.5\linewidth}
    \centering
    \begin{subfigure}[t]{2.8cm}
    \includegraphics[width=2.8cm]{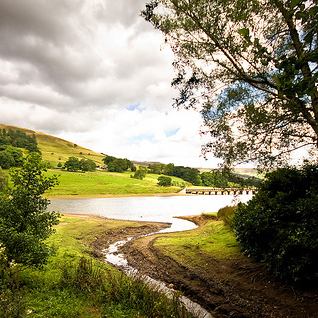}
    \caption{Clean image \\ \textbf{Softmax}: 0.967 \\\textbf{Ours}: 0.00519}\label{fig:comp_1}
    \end{subfigure}
    \begin{subfigure}[t]{2.8cm}
    \includegraphics[width=2.8cm]{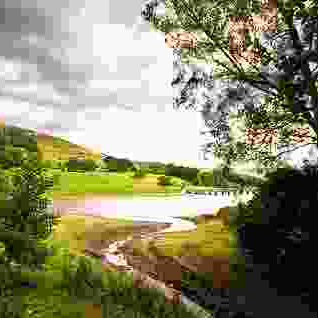}
    \caption{JPEG (Q = 5) \\ \textbf{Softmax}: 0.968 \\\textbf{Ours}: 0.00508}\label{fig:comp_2}
    \end{subfigure}
    \begin{subfigure}[t]{2.8cm}
    \includegraphics[width=2.8cm]{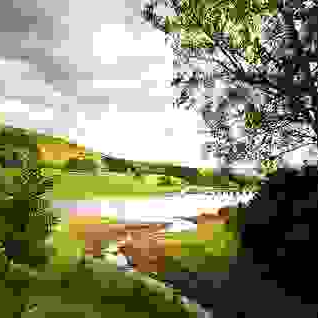}
    \caption{JPEG (Q = 1) \\ \textbf{Softmax}: 0.969 \\\textbf{Ours}: 0.00507}\label{fig:comp_3}
    \end{subfigure} 
  \end{minipage}
  \begin{minipage}[b]{0.5\linewidth}
    \centering
    \begin{subfigure}[t]{2.8cm}
    \includegraphics[width=2.8cm]{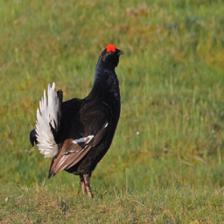}
    \caption{Clean Image \\ Label: Black grouse \\\textbf{Softmax}:0.291 \\\textbf{Ours}: 0.00283 }\label{fig:adv_1}
    \end{subfigure}
    \begin{subfigure}[t]{2.8cm}
    \includegraphics[width=2.8cm]{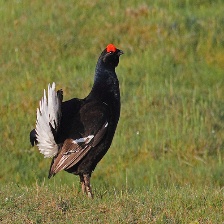}
    \caption{Adversarial Image \\ Label: Hen \\\textbf{Softmax}: 0.378 \\\textbf{Ours}: 0.00235}\label{fig:adv_2}
    \end{subfigure}
    \begin{subfigure}[t]{2.8cm}
    \includegraphics[width=2.8cm]{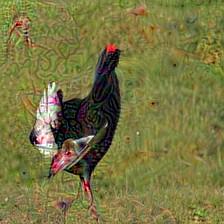}
    \caption{Adversarial Image \\ Label: Crane \\\textbf{Softmax}: 0.98 \\\textbf{Ours}: 0.0018}\label{fig:adv_3}
    \end{subfigure}
    
  \end{minipage} 
  \caption{Figures (a-c) show the effect of additive Gaussian noise, Figures (d-f) show the effect of blur, Figures (g-i) show JPEG Compression, while Figures (j-k) illustrate adversarial examples. In all cases, the quality of image decreases from left to right. We see that the proposed approach shows confidence(unnormalized) drop when distortions are increased, whereas softmax confidence does not exhibit this property.  }
  \label{fig:examples} 
\end{figure*}

\item 
\textbf{Gaussian Blurring:} Gaussian Blurring represents a filtering operation that removes high-frequency image content. When there is less evidence for the presence of important features, confidence of classification must decrease. While this indeed holds for Figure 2(b) for the case of MNIST, we do not see this behaviour for ImageNet (Figure 4(b)). For MNIST, we vary the standard deviation of the Gaussian kernel from 0 to 2, while for ImageNet it is varied from 0 to 36. 

\item 
\textbf{JPEG Compression:} Another important family of distortions that we often encounter is the loss of image content that occurs due to JPEG Compression. Such a compression is lossy in nature, and this loss is decided by the quality factor used in JPEG compression, which is varied typically from 20 to 100. In this case we expect to see a monotonically increasing profile w.r.t quality index, which both softmax and the proposed method achieve.

\item 
\textbf{Adversarial examples:} Adversarial images were generated according to the algorithm provided in \cite{moosavi2016deepfool}. We generated adversarial images for the entire validation set of ILSVRC12 dataset. After presenting both the original and adversarial images and computing confidences for both, we consider a rise in confidence for the adversarial case (when compared to the clean image) as a failure. We count the number of times both methods - softmax and the proposed approach - fail, and present the results in Table \ref{table:1}.
\end{itemize}

\begin{table}[h!]
\centering
\begin{tabular}{ |c|c|}
 \hline
 \textbf{$\#$ Softmax fails} &  \textbf{$\#$ Proposed measure fails} \\
 \hline
 5795  & 2214 \\
\hline
\end{tabular}

   \caption{Performance of confidence measures for adversarial examples. Adversarial Images were generated for the entire validation set of ImageNet using \cite{moosavi2016deepfool}.}
    \label{table:1}
\end{table}

\section{Discussion and Conclusion}
We have introduced a novel method of measuring the confidence of a neural network. We showed the sensitivity of softmax function to the scale of the input, and how that is an undesirable quality. The density modelling approach to confidence estimation is quite general - while we have used a Gaussian with diagonal covariance, it is possible to use much more sophisticated models for the same task. Our results show that in most cases the diagonal Gaussian works well, and mostly outperforms a softmax-based approach. We hypothesize that performance suffers in case of Gaussian blurring and partly in the case of Adversarial examples due to difficulties associated with high-dimensional density estimation. Future work looking at more sophisticated density models suited to high-dimensional inference are likely to work better.

\bibliographystyle{IEEEbib}
\bibliography{camera-ready_icme2017template}

\section*{Appendix}

Here we shall elucidate the proof of Lemma \ref{lemma}.

\begin{proof}
Consider $s_i(k z) = \frac{\exp(k z_i)}{\sum_j \exp(k z_j)}$. This can be re-written as follows.

\begin{align*}
    s_i(k z) & =  \frac{\exp(z_i) \exp((k-1)z_i)}{\sum_j \exp(k z_j)}\\
    & = \frac{\exp(z_i)}{\sum_j \frac{\exp(k z_j)}{\exp((k-1)z_i)}}\\
    & = \frac{\exp(z_i)}{\sum_j \exp(k z_j - (k-1) z_i)} 
\end{align*}

Comparing the denominator terms of the above expression and of $s_i(z)$, we arrive at the following condition for $s_i(k z) > s_i(z)$, assuming that each element $z_j$ is independent of the others. We shall complete the proof by contradiction. For this, let us assume $s_i(k z) < s_i(z)$. This implies the following.
\begin{align*}
    \exp(k z_j - (k-1) z_i) > \exp(z_j) ~~ \forall j \in [1,...N] \\
    \rightarrow (k-1) z_j > (k-1) z_i 
\end{align*}

The statement above is true iff exactly one of the two conditions hold:

\begin{itemize}
\item $k-1 < 0, \rightarrow k < 1$. This is false, since it is assumed that $k > 1$. 
\item $z_j > z_i$. This is false since $z_i$ is assumed to be the maximum of $z$.
\end{itemize}

We arrive at a contradiction, which shows that the premise, i.e.; $s_i(k z) < s_i(z)$ is false.

\end{proof}

\end{document}